\documentclass[journal,10pt]{IEEEtran}
\usepackage{graphicx}
\usepackage{amsmath}
\usepackage{array}
\usepackage{epsfig}
\usepackage{amssymb}
\usepackage{epstopdf}
\usepackage{float}
\usepackage{multirow}
\usepackage[small]{caption}
\usepackage{chngpage}

\usepackage{xspace}

\ifCLASSINFOpdf
\else
\fi

\usepackage{url}
\begin{document}

\title{Exploiting Depth from Single Monocular Images for Object Detection and Semantic Segmentation}

\author{Yuanzhouhan Cao, Chunhua Shen,  Heng Tao Shen %
	\thanks{Y. Cao, and C. Shen are with The University of Adelaide, Australia.}%
	\thanks{H. T. Shen is with the The University of Queensland, Australia.}%
}

\markboth{Appearing in IEEE Transactions on Image Processing, October 2016}%
{Cao et al.: Exploiting Depth from Single Monocular Images for Object Detection and Semantic Segmentation}

\maketitle

\begin{abstract}

Augmenting RGB data with measured depth has been shown to improve the performance of a range of tasks in computer vision including object detection and semantic segmentation. Although depth sensors such as the Microsoft Kinect have facilitated easy acquisition of such depth information, the vast majority of images used in vision tasks do not contain depth information. In this paper, we show that augmenting RGB images with estimated depth can also improve the accuracy of both object detection and semantic segmentation. Specifically, we first exploit the recent success of depth estimation from monocular images and learn a deep depth estimation model. Then we learn deep depth features from the estimated depth and combine with RGB features for object detection and semantic segmentation. Additionally, we propose an RGB-D semantic segmentation method which applies a multi-task training scheme: semantic label prediction and depth value regression. We test our methods on several datasets and demonstrate that incorporating information from estimated depth improves the performance of object detection and semantic segmentation remarkably.

\end{abstract}

\begin{IEEEkeywords}
Object detection, semantic segmentation, depth estimation, deep convolutional networks.
\end{IEEEkeywords}

\IEEEpeerreviewmaketitle

\tableofcontents
\clearpage

\section{Introduction}

Since the formulation of color images are two-dimensional projections of the three-dimensional world with depth information inevitably discarded, depth information and color information are complementary in restoring the real three-dimensional scenes. It is not difficult to conjecture that the performance of vision tasks such as object detection and semantic segmentation is likely to be improved if we can take advantage of the depth information.

Depth has been proven to be a very informative cue for image interpretation in a significant amount of work \cite{guptaECCV14,SongX14,bo_iros11}. In order to exploit depth information, most of these algorithms require depth data captured by depth sensors, as well as camera parameters, to relate point clouds to pixels. However, despite recent development of a range of depth sensors, most computer vision datasets such as the ImageNet and the PASCAL VOC are still RGB only. Moreover, images within these datasets are captured by different cameras with no camera parameters available, rendering it unclear how to generate accurate point clouds.

In this paper, we take advantage of deep convolutional neural networks (CNN) based depth estimation methods and show that the performance of object detection and semantic segmentation can be improved by incorporating an explicit depth estimation process. This may appear to be somewhat counter-intuitive, because the depth values estimated by CNN models are correlated to RGB images. The valuable depth information is acquired by depth sensors and lies in the RGB-D training data on which the depth estimation model is trained. Successful RGB-D object detection and semantic segmentation models need to be able to minimize the correlativity between RGB data and estimated depth and recover the depth information independently.

We here propose two methods to exploit depth information from color images for object detection and semantic segmentation. The two methods make use of estimated depth in different forms. The first method manages to learn deep depth features from the estimated depth. The learned depth feature is combined with RGB features for both object detection and semantic segmentation. The depth information and color information are fused at a later stage. For semantic segmentation, the estimated depth is used to compute an additional task loss. During training, the two losses jointly update the layers in an unified network in an end-to-end style. Thus, the depth information and color information are fused at an earlier stage.

Depth estimation from single monocular images is the first step of our proposed methods. Recently, CNNs have been applied to depth estimation and shown great success. For example, Eigen \textit{et al.} \cite{EigenPF14} proposed to estimate depth using multi-scale CNNs. It directly regresses on the depth using CNN with two components: one estimates the global structure of the scene while the other one refines the structure using local information. Here, we follow the most recent work of Liu \textit{et al.} \cite{LiuSLR15} which learns a deep convolutional neural fields (DCNF) model for depth estimation. It jointly learns the unary and pairwise potentials of conditional random fields (CRF) in a single CNN framework, and achieved state-of-the-art performance.

To summarize, we highlight the main contributions of this work as follows.
\begin{enumerate}
\item
For the tasks of object detection and semantic segmentation, we propose to incorporate estimated depth information with RGB data and improve the performance.

\item
We propose two methods of exploiting depth from single monocular images for object detection and semantic segmentation.

\item
We show that it is possible to improve the performance of object detection and semantic segmentation by exploiting relative data which do not share the same set of labels.
\end{enumerate}

The rest of the paper is organized as follows: In chapter \ref{chapt2}, we briefly review some related work. In chapter \ref{chapt3}, we introduce the DCNF model which we use for depth estimation. In chapter \ref{chapt4} and chapter \ref{chapt5}, we describe our RGB-D object detection and semantic segmentation frameworks respectively. Chapter \ref{chapt6} shows the experimental results and chapter \ref{chapt7} concludes the paper.

\section{Related Work}\label{chapt2}
Our work is inspired by the recent progresses of depth estimation, vision with RGB-D data, object detection and semantic segmentation. In this section, we briefly introduce the most related work.

\subsection{Depth Estimation}
Depth estimation from a single RGB image is the first step of our method.  As a part of the 3D structure understanding, traditional depth estimation methods are mainly based on geometric models. For example, the works in \cite{Hedau,NIPS2010_4120,SchwingECCV2012} rely on box-shaped models and fit the box edges to those observed in the image. These methods rely heavily on geometric assumptions and fail to provide a detailed 3D description of the scene. Other methods attempt to exploit additional information. In particular, the authors of \cite{RussellT09} estimate depth through user annotations. The work of \cite{LiuCVPR2010,Ladicky_2014_CVPR} make use of semantic class labels. Given the fact that the extra source of information is not always available, most of recent works formulate depth estimation as a Markov Random Fields (MRF) \cite{Saxena2005,Saxena2009,Saxena073-ddepth} or Conditional Random Fields (CRF) \cite{Liu_2014_CVPR} learning problem. These methods learn the parameters of MRF/CRF from a training set of monocular images and their ground-truth depth maps in a supervised fashion. The depth estimation problem is then formulated as maximum a posteriori (MAP) inference within the CRF model.

Most of the aforementioned algorithms use hand-crafted features such as texton, GIST, SIFT, PHOG, etc.
 With the fast development of DCNN recently, some works attempt to solve the depth estimation problem in a deep network and have achieved very impressive performance \cite{EigenPF14,LiuSLR15,Li_2015_CVPR}. In this article, we follow the deep conditional neural fields (DCNF) model introduced in \cite{LiuSLR15} for depth estimation.

\subsection{Incorporating Depth}
These methods can be roughly divided into two categories. The first type attempts to recover the real 3D shape of the scenes and explores 3D feature descriptors. To name a few,  Song \textit{et al.} \cite{SongX14} extended the deformable part based model (DPM) from 2D to 3D by proposing four point-cloud based shape features: point density feature, 3D shape feature, 3D normal feature and Truncated Signed Distance Function (TSDF) feature.  Bo \textit{et al.} \cite{bo_iros11} developed a set of kernel features on depth images that model the size, 3D shape, and depth edges for object recognition. In \cite{540}, 3D information of object parts is treated as constrains for object detection. Sun \textit{et al.} \cite{MinECCV10} also estimate 3D shape of objects to improve object detection accuracy. Other algorithms make use of the context information such as inter-object relations or object-background relations \cite{citeulike:3731461,457,Lin_2013_ICCV}.  These methods are able to provide a multi-view understanding of objects but also need large amounts of 3D shape training data.

The second category encodes the depth values as a 2D image and combines with the RGB image to formulate the 2.5D data. For example,  Gupta \textit{et al.} \cite{guptaECCV14} proposed a depth map embedding scheme that encodes the height above ground, angle with gravity and horizontal disparity (HHA) as an additional input to RGB for object detection and semantic segmentation. Janoch \textit{et al.} \cite{JanochKJBFSD11} extracted HOG features from depth images and trained a DPM model on these depth features for object detection. Schwarz \textit{et al.} \cite{SchwarzSB15} proposed an object-centered colorization scheme which is tailored for object recognition and pose estimation. All these methods require direct measurements of depths.

\subsection{Object Detection}
Object detection has been an active research topic for decades. Conventional algorithms exploit hand-crafted feature descriptors such as SIFT \cite{DBLP:Lowe04}, HOG \cite{DT05,Felzenszwalb}, and generate mid-level image representations through Bag-of-Visual-Words (BoVW) \cite{Csurka04,LSP06,Sivic03,Zhang07localfeatures} or Fisher Vector (FV) \cite{PSM10,NIPS2013_4926} encoding schemes. In recent years, CNN based methods have been demonstrated to outperform all hand-crafted features based algorithms. In \cite{girshick2014rcnn}, Girshick \textit{et al.} combined regional proposals with CNN features and have achieved outstanding detection results. In their following work \cite{girshickICCV15fastrcnn}, the authors proposed an region of interest (RoI) pooling layer and further improve the processing speed. In \cite{Zhang_2015_CVPR}, Zhang \textit{et al.} improve the CNN based object detection with Bayesian optimization and structured prediction. In \cite{Ouyang_2015_CVPR}, Ouyang \textit{et al.} proposed a deformable CNN for object detection. It effectively integrates feature representation learning, part deformable learning, context modeling, model averaging and bounding box location refinement into the detection system.

\subsection{Semantic Segmentation}
Convolutional neural networks have also shown great potential in semantic segmentation \cite{pami_FarabetCNL13,NingDLPBB05,pinheiro_2014}. Recently, Long \textit{et al.} \cite{long_shelhamer_fcn} proposed a fully convolutional network (FCN) for semantic segmentation. It is the first work to train FCN end-to-end for pixelwise prediction. Many recent works combine DCNNs and CRFs for semantic segmentation and have achieved state-of-the-art performance. Schwing \textit{et al.} \cite{SchwingU15} jointly learned the dense CRFs and CNNs. The pairwise potential functions only enforce smoothness and are not CNN-based. Lin \textit{et al.} \cite{LinSRH15} jointly trained FCNs and CRFs and learn CNN-based general pairwise potential functions.

Unlike the aforementioned methods which are purely based on color information, our methods make use of depth information to improve objected detection and semantic segmentation performance.

\section{Depth Estimation Model}\label{chapt3}
We use the DCNF model introduced by Liu \textit{et al.} \cite{LiuSLR15} for depth estimation. It jointly learns the unary term and pairwise term of continuous CRF in an unified network. In this section, we first introduce continuous CRF model, then we introduce the DNCF model with fully convolutional networks and superpixel pooling.

\subsection{Continuous CRF}
Similar to \cite{LiuCVPR2010,Saxena2009,Liu_2014_CVPR}, the images are represented as sets of small homogeneous regions (superpixels). The depth estimation is based on the assumption that pixels within a same superpixel have similar depth values. Each superpixel is represented by the depth value of its centroid. Let $\mathbf{x}$ be an image and $\mathbf{y}=[y_1,...,y_n]^\top\in\mathbb{R}^{n}$  be a vector of superpixels in $\mathbf{x}$, the conditional probability distribution of the data could be modelled as the following density function:
\begin{equation}
\mathrm{Pr}(\mathbf{y}|\mathbf{x})=\frac{1}{\mathrm{Z}(\mathbf{x})}\exp(-\mathrm{E}(\mathbf{y},\mathbf{x})),
\end{equation}
where $\mathrm{E}$ is the energy function and $\mathrm{Z}$ is the partition function:
\begin{equation}
\mathrm{Z}(\mathbf{x})=\int_\mathbf{y}\exp\{-\mathrm{E}(\mathbf{y},\mathbf{x})\}\mathrm{d}_\mathbf{y}.
\end{equation}
Since the depth value $\mathbf{y}$ is continuous, there are no any approximation methods here. The depth value of a new image could be predicted through the following MAP inference:
\begin{equation}
\mathbf{y}^*=\underset{\mathbf{y}}{\mathrm{argmax}}\mathrm{Pr}(\mathbf{y}|\mathbf{x}).
\end{equation}
The energy function is formulated as a typical combination of unary potentials $\mathrm{U}$ and pairwise potentials $\mathrm{V}$ over the nodes (superpixels) $\mathcal{N}$ and edges $\mathcal{S}$ of the image $\mathbf{x}$:
\begin{equation}
\mathrm{E}(\mathbf{y},\mathbf{x})=\sum_{p\in{\mathcal{N}}}\mathrm{U}(y_p,\mathbf{x})+\sum_{(p,q)\in{\mathcal{S}}}\mathrm{V}(y_p,y_q,\mathbf{x}).
\end{equation}

\subsubsection{Unary potential function}
The unary term $\mathrm{U}$ regresses the depth value from a single superpixel. It is formulated by the following least square loss:
\begin{equation}
\mathrm{U}(y_p,\mathbf{x};\boldsymbol{\theta})=(y_p-z_p(\boldsymbol{\theta}))^2,\forall{p}=1,\dots,n;
\end{equation}
where $z_p$ is the regressed depth value of superpixel $p$ parametrized by parameters $\boldsymbol{\theta}$.
\subsubsection{Pairwise potential function}
The pairwise term $\mathrm{V}$ encourages neighbouring superpixels with similar appearances to have similar depth values. Is is constructed by 3 types of similarities: color, color histogram and texture disparity in terms of local binary patterns (LBP), which are represented as follows:
\begin{equation}
\mathrm{S}_{pq}^{(k)}=\exp\lbrace-\gamma\|s_{p}^{(k)}-s_{q}^{(k)}\|\rbrace, k=1,2,3;
\end{equation}
where $s_{p}^{(k)}$ and $s_{q}^{(k)}$ are the observation values of superpixel $p$ and $q$ from color, color histogram and LBP respectively. $\|\cdot\|$ denotes the $\ell_2$ norm of a vector and $\gamma$ is a constant.

The similarity observation values are fed into a fully-connected layer:
\begin{equation}
\mathrm{R}_{pq}=\boldsymbol{\beta}^{\top}[\mathrm{S}_{pq}^{(1)},\dots,\mathrm{S}_{pq}^{(k)}]^{\top}=\sum_{k=1}^\mathrm{K}\beta_k\mathrm{S}_{pq}^{(k)},
\end{equation}
where $\boldsymbol{\beta}=[\beta_1,\dots,\beta_k]^{\top}$ are trainable parameters. Finally, the pairwise potential function is formulated as:
\begin{equation}
\mathrm{V}(y_p,y_q,\mathbf{x};\boldsymbol{\beta})=\frac{1}{2}\mathrm{R}_{pq}(y_p-y_q)^2.
\end{equation}

\subsection{Deep Convolution Neural Field Model}
The whole network is composed of a unary part, a pairwise part and a CRF loss layer. During training, an input image is first over-segmented into $n$ superpixels. Then the entire image is fed into the unary part and outputs an $n$-dimensional vector containing regressed depth values of the $n$ superpixels. Specifically, the unary part is composed of a fully convolution part and a superpixel pooling part. After fully convolution, an input image is convolved into a set of convolutional feature maps. The superpixel pooling part takes the convolutional feature maps as the input and outputs $n$ superpixel feature vectors. The $n$ superpixel feature vectors are then fed into 3 fully-connected layers to produce the unary output.

\begin{figure*}
	\begin{center}
		\includegraphics[scale=.92862]{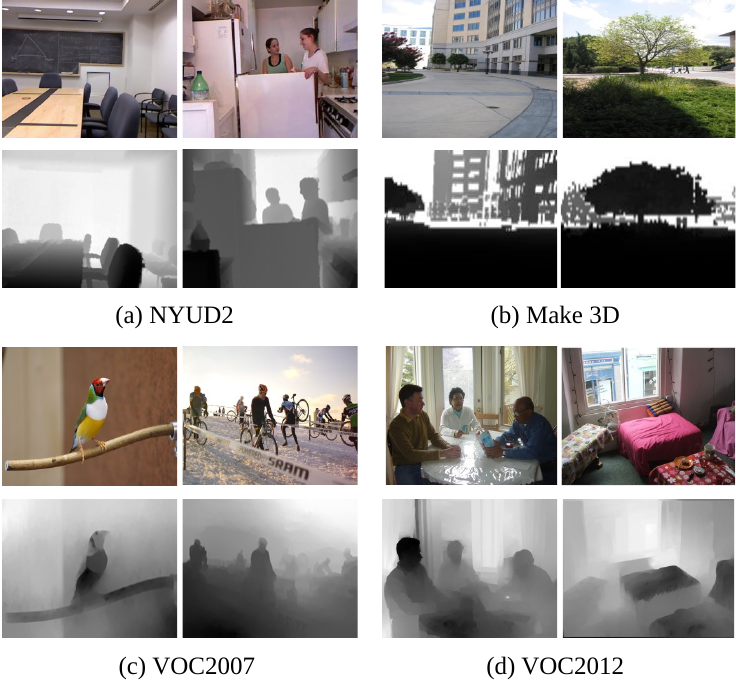}
	\end{center}
	\caption{Examples of ground-truth depth maps and estimated depth maps. (a) NYUD2 dataset (ground-truth) (b) Make3D dataset (ground-truth) (c) VOC2007 dataset (estimated) (d) VOC2012 dataset (estimated).}
	\label{fig:estD-example}
\end{figure*}

The pairwise part takes similarity vectors (each with 3 components) of all neighbouring superpixel pairs as input and feeds each of them into a fully-connected layer (parameters are shared among different pairs), then outputs a vector containing all the 1-dimensional similarities for each of the neighbouring superpixel pairs. The continuous CRF loss layer takes the outputs from the unary and the pairwise terms to minimize the negative log-likelihood. Fig. \ref{fig:estD-example} shows some examples of ground-truth depth maps and depth maps estimated by the DCNF model.

\section{RGB-D Object Detection}\label{chapt4}
We now describe how we exploit depth information from RGB images for object detection. Specifically, we build our models upon the R-CNN \cite{girshick2014rcnn} and Fast R-CNN \cite{girshickICCV15fastrcnn} models and propose to learn depth features for each object proposal. We use the extracted depth features as extra information for object detection.

\begin{figure}
	\begin{center}
		\includegraphics[scale=.59]{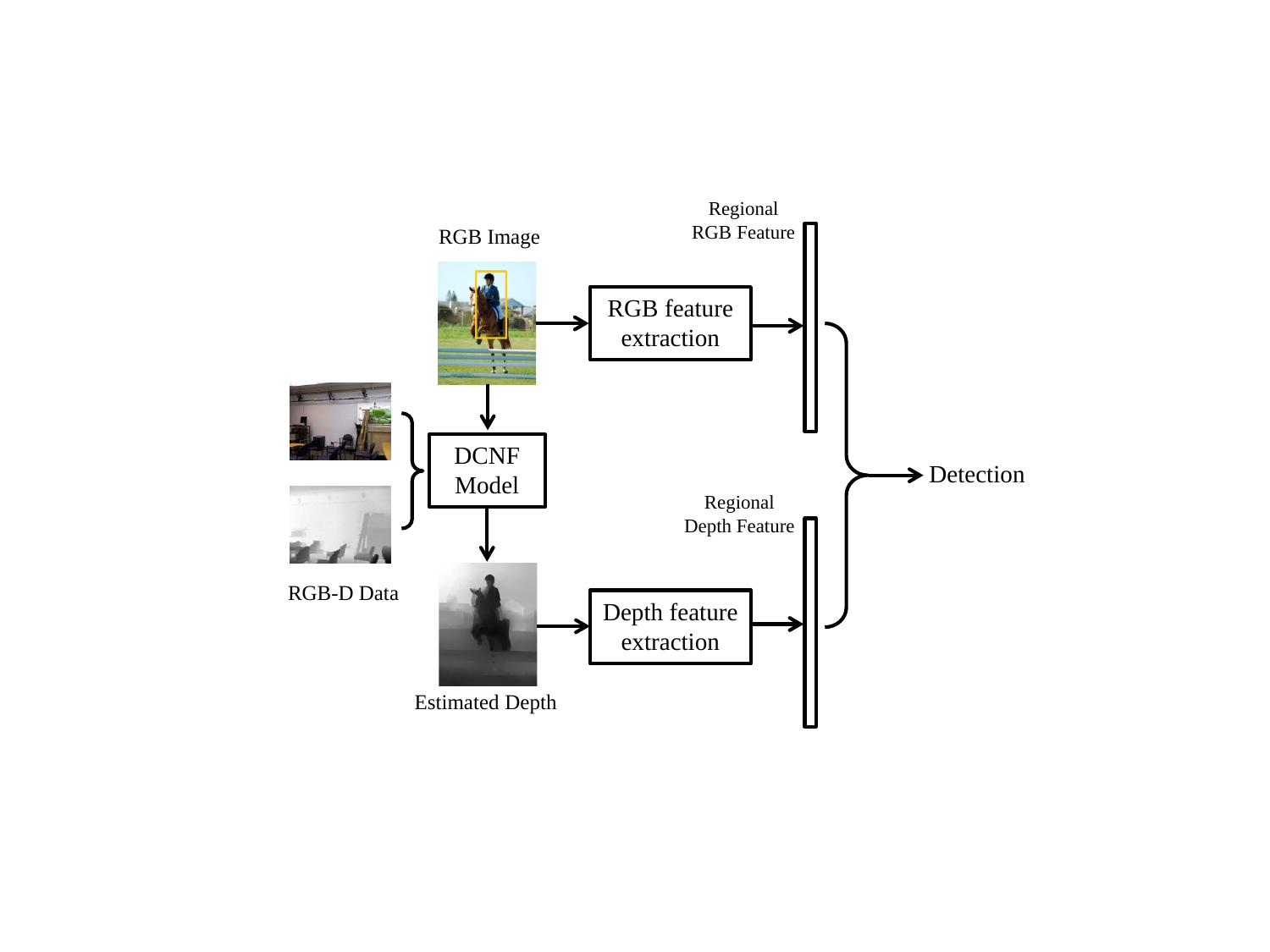}
	\end{center}
	\caption{An overview of our RGB-D detection system. A DCNF model is first learned from RGB-D datasets. The input RGB image and its estimated depth image are fed into two feature extraction networks. The RGB feature and depth feature of each object proposal are concatenated for object detection.}
	\label{fig:rgbd-detect}
\end{figure}

\begin{figure*}
	\begin{center}
		\includegraphics[scale=.70]{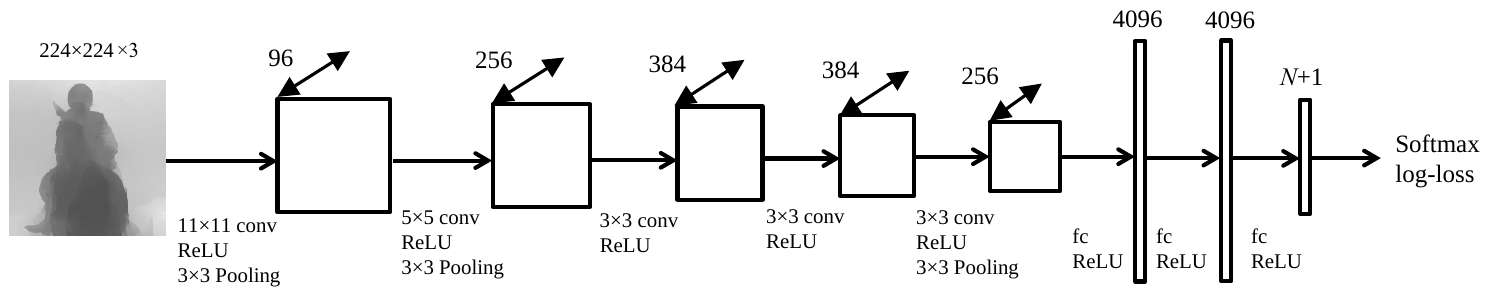}
	\end{center}
	\caption{Detailed structure of the depth feature learning network of our RGB-D R-CNN detector. It consists of five convolutional layers and three fully-connected layers. During training, the input image patch is first resized to $224 \times 224 \times 3$ and fed into the network. The last layer calculates softmax log-loss and back-propagates through the network for parameter updating. We use the output of the first or second fully-connected layer which is a 4096-dimensional vector as the additional depth feature for object detection.}
	\label{fig:rgbd-detect-net}
\end{figure*}

\subsection{System Overview}
We design our RGB-D detection system based on two observations. Firstly, pixels within a single  object have similar depth values. Secondly, the camera and the depth sensor have the same viewing angles in generating RGB-D datasets such as NYUD2 and Make3D.  As a result, an object in a depth image exhibits the same 2D shapes with its RGB counterpart, only with RGB values replaced by depth values (similar to intensity). Depth images can aggregate inter-object variations and eliminate intra-object variations. This is important for scene understanding, e.g., a chair appears in a painting or TV monitor should not be detected.

An overall structure of our RGB-D detection system is illustrated in Fig. \ref{fig:rgbd-detect}. As we can see from the figure, it comprises 3 parts: depth estimation and encoding, RGB and depth feature extraction and object detection. During testing, we first estimate depth values of the input RGB image using the DCNF model trained on RGB-D dataset. With the estimated depth, we encode the estimated depth values into a 3-channel image. Similar to R-CNN and Fast R-CNN, we also follow the ``recognition using regions" paradigm and generate object proposals on RGB images. We extract RGB and depth features of each object proposal through two separate streams. The RGB features capture intra-object information such as color and texture while the depth features aggregate 2D shapes of objects. Finally, we concatenate the two features for object detection.

\subsection{RGB-D R-CNN Object Detector}\label{analy}
The training of the R-CNN detector is a multi-stage pipeline. It first fine-tunes a CNN network pre-trained on a large classification dataset such as ImageNet. Then SVMs are trained on features extracted from the fine-tuned network in the first stage. These SVMs act as detectors.

\subsubsection{Depth encoding and feature learning}
The output of the DCNF model is a single-channel depth map. In order to be used as the input of our depth feature learning network, we log-normalize these depth values to the range of $[0,255]$ and duplicate into three channels. Here we do not apply any geometric contour cues. This is mainly because the calculation of geometric cues (normals, normal gradients, depth gradients, etc.) requires point clouds as well as accurate depth values. Since we use estimated depth values, the outliers could affect the accuracy of geometric cues. Moreover, we do not access camera parameters for most of the color images which makes it impractical to recover the point clouds.

The CNN pre-trained on a large classification dataset can be used as a generic extractor of mid-level features \cite{Oquab2014}. Since RGB and depth images also share similar structures, we fine-tune a CNN pretrained on RGB images using depth images for depth feature learning. The structure of our regional depth feature learning network is illustrated in Fig. \ref{fig:rgbd-detect-net}. It has 5 convolutional layers and 3 fully-connected layers interlaced with ReLU and max pooling layers. The last layer of the network is a classification layer with $N+1$ channels, where $N$ is the number of classes and the added 1 for background. We initialize the parameters of our network with AlexNet \cite{NIPS2012_4824}. During fine-tuning, each object proposal is resized to 224 $\times$ 224 $\times$ 3 and then fed into the network. For each class, we use object proposals that have $>50\%$ intersection over union with the ground-truth boxes as positive training samples and the rest as negative training samples.  During testing, we use the output of the fully-connected layers as depth feature.

\subsubsection{Detector learning}
The learnt depth features are 4096-dimensional feature vectors. We fine-tune another network using RGB images for RGB feature learning. The depth and RGB feature learning networks have the same structure but do not share parameters. For each of the object proposals, we concatenate the depth feature and RGB feature to be a 8192-dimensional feature vector. Since the RGB and depth images have the same scales and the two feature learning networks have the same structures, we do not normalize the features after concatenation.

After feature concatenation we train a binary SVM for each class as object detectors. Training is done using liblinear \cite{jmlr_FanCHWL08} with hyper-parameters $C=0.001, B=10, w_1=2$ where $C$ is the SVM trade-off parameter, $B$ is bias term and $w_1$ is the cost factor on hinge loss for positive examples. We use the same SVM hyper-parameters as in \cite{girshick2014rcnn} and find that the final detection performance is not sensitive to these parameters. Following \cite{girshick2014rcnn}, the positive examples are from the ground truth boxes for the target class and the negative examples are defined as regions having $<30\%$ intersection over union with the ground truth boxes. During training, we adopt standard hard negative mining. Hard negative mining converges quickly and in practice detection accuracy stops increasing after only a single pass over all images.

\begin{figure*}
	\begin{center}
		\includegraphics[scale=.67]{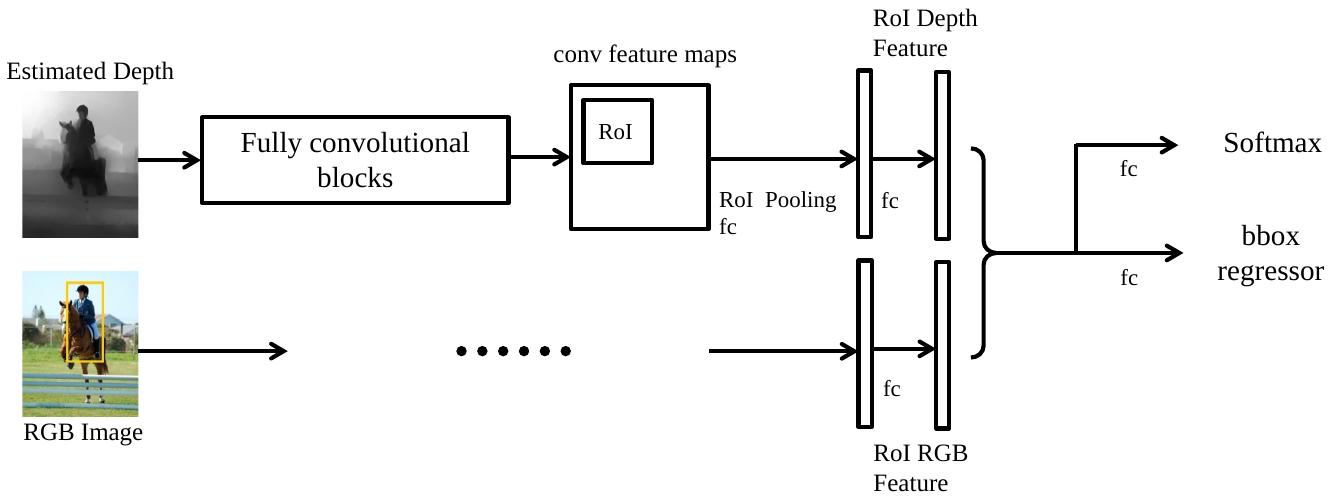}
	\end{center}
	\caption{Detailed structure of our RGB-D Fast R-CNN network. It inputs the entire RGB and depth images and outputs two convolutional feature maps. After RoI pooling and fully-connected layers, we concatenate the RGB and depth feature of each RoI and calculate a multi-task loss. The RGB and depth feature extraction network have the same structure before concatenation. We omit the RGB feature extraction stream in the figure.}
	\label{fig:frcnn-detect-net}
\end{figure*}

\begin{figure*}
	\begin{center}
		\includegraphics[scale=.61]{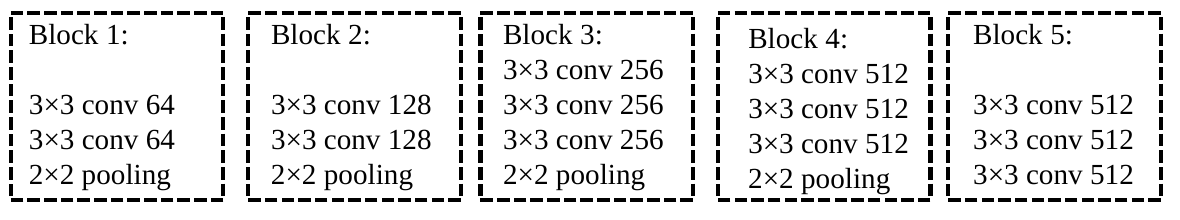}
	\end{center}
	\caption{Detailed structure of the fully convolutional blocks in Figs. \ref{fig:frcnn-detect-net}, \ref{fig:rgbd-seg}, \ref{fig:concat-seg}. The parameters are transferred from VGG16.}
	\label{fig:vgg16-detail}
\end{figure*}

\subsection{RGB-D Fast R-CNN Object Detector}
As mentioned above, the training of our RGB-D R-CNN detector is a multi-stage pipeline. For the SVM training, both depth and RGB features are extracted from each object proposal in each image. As a result, training of R-CNN detector is expensive in space and time. Fast R-CNN detector alleviated these disadvantages. It is based on fully convolutional networks which take as inputs arbitrarily sized images, and output convolutional spatial maps. Hence we propose another RGB-D detection network based on Fast R-CNN.

\subsubsection{Architecture}
We show the architecture of our RGB-D Fast R-CNN network in Fig. \ref{fig:frcnn-detect-net}. During training, the network takes as the input the entire RGB and depth images and a set of object proposals to produce two convolutional feature maps, one for RGB and one for depth. Then for each object proposal a region of interest (RoI) pooling layer extracts a fixed-length feature vector from the feature map. The RGB and depth feature vectors of each RoI are concatenated and fed into a sequence of fully connected layers that finally branch into two output layers: one that produces softmax probability estimates over $N+1$ classes (1 for background) and another that outputs four real-valued numbers for each of the $N$ object classes. The detailed structure of the fully convolutional blocks is illustrated in Fig. \ref{fig:vgg16-detail}. We initialize our network with the VGG-16 net \cite{Simonyan14c}. Before the feature concatenation, The RGB and depth streams in our RGB-D Fast R-CNN network have the same structure but do not share parameters.

\subsubsection{RoI Pooling and multi-task loss}
An RoI is a rectangular window on a convolutional feature map. Each RoI is defined by a four-tuple $(r,c,h,w)$ that specifies its top-left corner $(r,c)$ and its height and width $(h,w)$. The RoI pooling layer uses max pooling to convert the features inside any valid region of interest into a small feature map with a fixed spatial extend of $H \times W$, where $H$ and $W$ are layer hyper-parameters that are independent of any particular RoI.

During training, each RoI is labelled with a ground-truth class $u$ and a ground-truth bounding-box regression target $v$. We use a multi-task loss $L$ on each labelled RoI to joint train for classification and bounding box regression:
\begin{equation}
\mathrm{L}(p,u,t^u,v)=L_{cls}(p,u)+\lambda[u\geq1] \mathrm{L}_{loc}(t^u,v),
\end{equation}
where $L_{cls}(p,u)=-\log p_u$ is log loss for true class $u$. $[u\geq1]$ evaluates to 1 when $u\geq1$ and 0 otherwise. $L_{loc}$ is defined over a tuple of true bounding-box regression targets for class $u,v = (v_x,v_y,v_w,v_h)$, and a predicted tuple $t^u = (t_{x}^u,t_{y}^u,t_{w}^u,t_{h}^u)$, again for class $u$. For bounding-box regression, the loss is:
\begin{equation}
\mathrm{L}_{loc}(t^u,v)=\sum_{i\in{\{x,y,w,h\}}}\mathrm{smooth}_{\mathrm{L_1}}(t_i^u-v_i),
\end{equation}
in which
\begin{equation}
\mathrm{smooth}_{\mathrm{L_1}}(t_i^u-v_i)=
\begin{cases}
0.5x^2&\mathrm{if}  |x|<1;\\
|x|-0.5&\mathrm{otherwise}.
\end{cases}
\end{equation}
\section{RGB-D Semantic Segmentation}\label{chapt5}
In this section, we describe how we exploit depth for semantic segmentation. We first elaborate our RGB-D semantic segmentation with feature concatenation. Then we show the RGB-D semantic segmentation method which applies multi-task joint training.

\subsection{RGB-D Semantic Segmentation by Feature Concatenation}
\begin{figure*}
	\begin{center}
		\includegraphics[scale=.62]{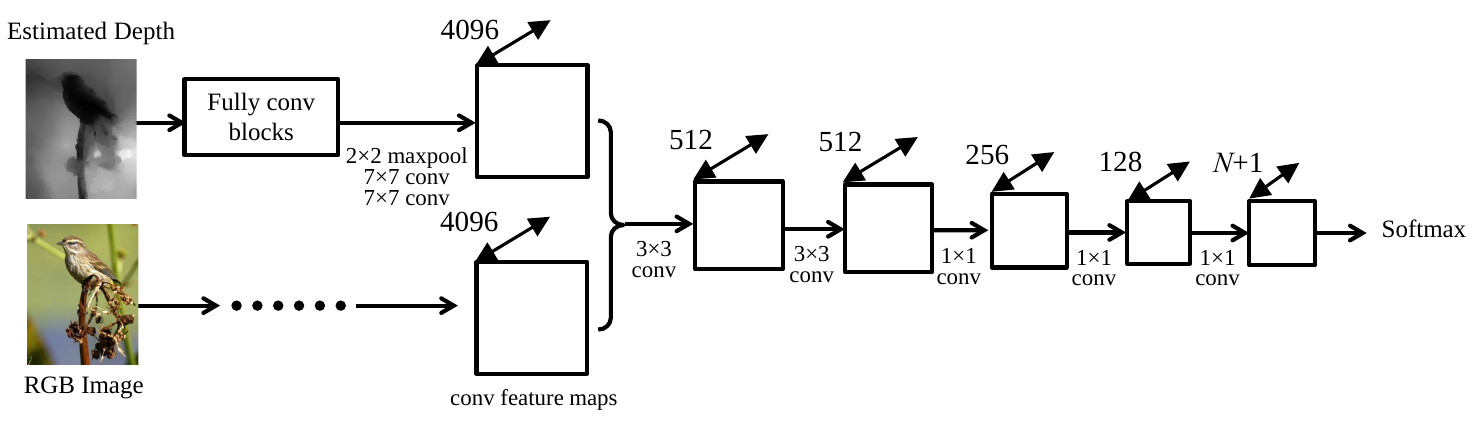}
	\end{center}
	\caption{Network structure of our RGB-D segmentation with feature concatenation. The network takes the entire RGB and depth images as input and outputs two feature maps. We concatenate the RGB and depth feature maps and calculate softmax log-loss after several convolutional layers.}
	\label{fig:concat-seg}
\end{figure*}

Similar to our RGB-D detection methods, we extract RGB and depth features from RGB and depth images respectively and concatenate the features for semantic segmentation. Specifically, we follow the fully convolutional networks \cite{long_shelhamer_fcn} which can take as inputs arbitrarily sized images. We show the network architecture in Fig. \ref{fig:concat-seg}. After depth estimation, the RGB and estimated depth images are fed into two separate fully convolutional processing streams and output two convolutional feature maps. The two feature maps are concatenated and fed into 5 convolutional layers with channels $512,512,256,128$ and $N+1$ respectively where $N$ is the number of classes and added 1 for background. Finally, a softmax layer is added during training to backpropagate the classification loss. We initialize our RGB-D segmentation network with the VGG-16 net \cite{Simonyan14c} and the last 5 convolution layers are added. Before feature map concatenation, the RGB and depth processing streams have the same network structure but do not share parameters.

\begin{figure*}
	\begin{center}
		\includegraphics[scale=.61]{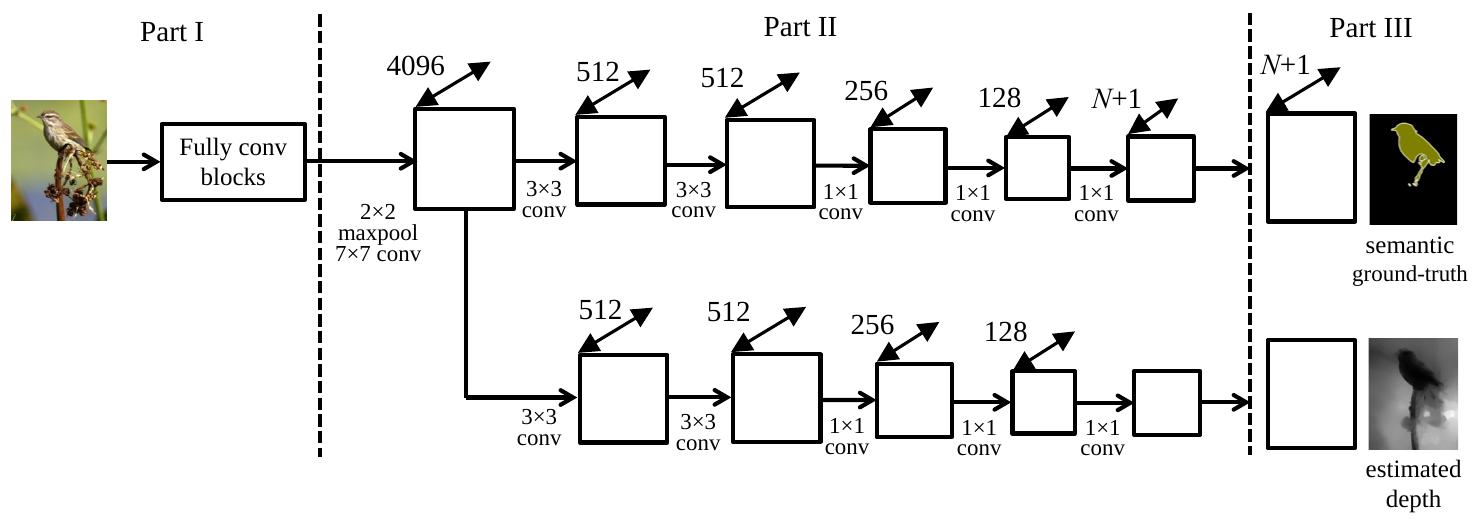}
	\end{center}
	\caption{An overview of our RGB-D segmentation by multi-task training. During training, an entire RGB  image is fed into the network and two loss functions are computed at the end of the network: softmax log-loss and regression loss. The two losses jointly update the network parameters in an end-to-end style. The detail of fully convolutional blocks is illustrated in Fig. \ref{fig:vgg16-detail}.}
	\label{fig:rgbd-seg}
\end{figure*}

\subsection{RGB-D Semantic Segmentation by Multi-task training}
The aforementioned method for RGB-D semantic segmentation has two separate networks for RGB and depth feature extraction. The two networks do not share parameters and thus are inefficient to train. Inspired by the Fast R-CNN detector which applies a multi-task training scheme, we propose another RGB-D semantic segmentation method. Since an RGB image has two labels after depth estimation: one semantic label and one depth label, we apply a multi-task loss which can jointly update a unified network in an end-to-end style.

\subsubsection{Network architecture}
The architecture of our multi-task RGB-D segmentation network is illustrated in Fig. \ref{fig:rgbd-seg}. As we can see from the figure, the network can be broadly divided into 3 parts. The first part takes as input the entire RGB image and output convolutional feature maps. We initialize this part with the layers up to ``fc6" in the VGG-16 net. During backpropagation, the semantic and depth loss share parameters in this part.

The second part has two sibling processing streams: one for color information and one for depth information. The color information processing stream contains 5 convolutional layers with channels $512,512,256,128$ and $N+1$ respectively. The depth information processing stream also contains 5 convolutional layers. The only difference is that the last convolutional layer only has 1 channel due to different loss function. The outputs of the second part are two feature maps of $N+1$ channels and 1 channel respectively. The multi-task loss part take these two feature maps as input and calculate two losses: semantic label prediction loss and depth value regression loss. The two processing streams in the second part backpropagate the two losses through the entire network.

\subsubsection{Multi-task loss}
The last part of our segmentation network calculates a multi-task loss: softmax log-loss for semantic label prediction, and least squared loss for depth regression. The two losses jointly update a unified network in an end-to-end style.

Let $\mathbf{M}$ be the output feature map of color information processing stream in second part of our segmentation network. We first upscale $\mathbf{M}$ to be the same size of original input image using nearest neighbour interpolation. Assuming the size of original input image is $h\times w$, and $d$ is the number of channels of the output feature map. The softmax log-loss computes

\begin{equation}
\mathrm{L_{color}}=-\sum_{i,j}(\mathrm{M}_{ijc}-\mathrm{log}\sum_{k=1}^\mathrm{d}\mathrm{e}^{\mathrm{M}_{ijk}}),
\end{equation}
where $i\in[1,h]$, $j\in[1,w]$, $k\in[1,d]$ and $c$ is the ground-truth label.

The estimated depth values are used to compute the depth regression loss. Let $\mathbf{F}$ be the output feature map of depth processing stream in the second part of our segmentation network. Similar to the softmax log-loss, we upscale $\mathbf{F}$ to be the same size of original input image $h\times w$. The least squared loss computes:

\begin{equation}
\mathrm{L_{depth}}=\sum_{i,j}(\mathrm{F}_{ij}-\mathrm{z}_{ij})^2,
\end{equation}
where $i\in[1,h]$, $j\in[1,w]$ and $z$ is the estimated depth values by DCNF model. Notably, different from DCNF model training, here we do not compute the least squared loss based on superpixels.

During backpropagation, the two separate losses are combined:

\begin{equation}
\mathrm{L}=\mathrm{L_{color}}+\lambda \mathrm{L_{depth}},
\end{equation}
where $\lambda$ is a hyper-parameter to control the balance between depth information and color information in network updation.

\section{Experiments}\label{chapt6}
In this section, we show the performance improvement of object detection and semantic segmentation introduced by exploited depth. Our experiments are organized in two parts: RGB-D object detection and RGB-D semantic segmentation. We test the RGB-D object detection on 4 datasets: NYUD2 \cite{Silberman:ECCV12}, B3DO \cite{JanochKJBFSD11}, PASCAL VOC 2007 and 2012 and report the mean average precision (mAP). During testing, a detection is considered correct if the intersection-over-union (IoU) between the bounding box and ground truth box is greater than 50\%. We test our RGB-D semantic segmentation on the VOC2012 dataset. The performance is measured by the IoU score.

\begin{table}
	\caption{Detection results of our RGB-D R-CNN detector on indoor objects of the VOC2012 dataset. The first row shows the results of RGB features only, the second row shows the results of estimated depth features only, the third row shows the results of combining RGB features and depth features.}
	\centering
	\label{table:VOC12_estD_results}
	\begin{tabular}{@{\hskip 0.1cm}c@{\hskip 0.2cm}c@{\hskip 0.2cm}c@{\hskip 0.2cm}c@{\hskip 0.2cm}c@{\hskip 0.2cm}c@{\hskip 0.2cm}c@{\hskip 0.2cm}c}
		\hline\noalign{\smallskip}
		                 & bottle   & chair    & table    & plant    & sofa     & tv       & mAP      \\
		\noalign{\smallskip}
		\hline
		\noalign{\smallskip}
		RGB              & 20.0     & 16.4     & 23.2     & 21.6     & 24.8     & 55.1     & 26.9     \\
		Estimated Depth  & 11.4     & 11.0     & 16.0     & 4.7      & 20.2     & 32.7     & 16.0     \\
		RGB-D (Estimated) & \bf 21.5 & \bf 20.8 & \bf 29.6 & \bf 24.1 & \bf 31.7 & \bf 55.1 & \bf 30.5 \\
		\hline
	\end{tabular}
\end{table}

For depth prediction of indoor datasets such as the NYUD2 and B3DO, we train the DCNF model on the NYUD2 dataset. For the VOC2007 and VOC2012 datasets which contain both indoor and outdoor scenes, we train the DCNF model on images from both the NYUD2 and Make3D \cite{Saxena073-ddepth} datasets.

\begin{table*}
	\renewcommand\arraystretch{1.25}
	\newcommand{\tabincell}[2]{\begin{tabular}{@{}#1@{}}#2\end{tabular}}
	\caption{Detection results on the VOC 2007 dataset of our RGB-D R-CNN detector. The first row shows the results of RGB features only, the second row shows the results of depth features only, the third row shows the results of combined RGB and depth features. The first three rows are results of RGB features directly extracted from AlexNet and the last two rows are the results of RGB features extracted from fine-tuned AlexNet.}
	\centering
	\label{table:07_estD_results}
	\begin{tabular}{@{\hskip 0.8mm}c@{\hskip 0.8mm} @{\hskip 0.8mm}c@{\hskip 0.8mm} @{\hskip 0.8mm}c@{\hskip 0.8mm} @{\hskip 0.8mm}c@{\hskip 0.8mm} @{\hskip 0.8mm}c@{\hskip 0.8mm} @{\hskip 0.8mm}c@{\hskip 0.8mm} @{\hskip 0.8mm}c@{\hskip 0.8mm} @{\hskip 0.8mm}c@{\hskip 0.8mm} @{\hskip 0.8mm}c@{\hskip 0.8mm} @{\hskip 0.8mm}c@{\hskip 0.8mm} @{\hskip 0.8mm}c@{\hskip 0.8mm} @{\hskip 0.8mm}c@{\hskip 0.8mm} @{\hskip 0.8mm}c@{\hskip 0.8mm} @{\hskip 0.8mm}c@{\hskip 0.8mm} @{\hskip 0.8mm}c@{\hskip 0.8mm} @{\hskip 0.8mm}c@{\hskip 0.8mm} @{\hskip 0.8mm}c@{\hskip 0.8mm} @{\hskip 0.8mm}c@{\hskip 0.8mm} @{\hskip 0.8mm}c@{\hskip 0.8mm} @{\hskip 0.8mm}c@{\hskip 0.8mm} @{\hskip 0.8mm}c@{\hskip 0.8mm} @{\hskip 0.8mm}c@{\hskip 0.8mm}}
		\noalign{\smallskip}
		\hline\noalign{\smallskip}
		                 &\small aero    &\small bike  &\small bird   &\small boat   &\small bottle  &\small bus &\small car    &\small cat    &\small chair   &\small cow    &\small table  &\small dog  &\small horse   &\small mbike  &\small person &\small plant &\small sheep  &\small sofa &\small train &\small tv &\small mAP \\
		\noalign{\smallskip}
		\hline
		\noalign{\smallskip}
		\small RGB \cite{girshick2014rcnn} & \small 59.3 & \small 61.8     & \small 43.1   & \small 34.0     & \small 25.1     & \small 53.1   &\small 60.6   &\small 52.8  &\small 21.7  &\small 47.8     &\small 42.7     &\small 47.8     &\small 52.5     &\small 58.5     &\small 44.6     &\small 25.6     &\small 48.3     &\small 34.0 &\small 53.1 &\small 58.0 &\small 46.2\\
		\small Depth & \small 39.1     & \small 33.7     & \small 21.2     & \small 15.6     & \small 10.9     & \small 36.1     &\small 43.2  &\small 30.7  &\small 12.0  &\small 21.3 &\small 27.7     &\small 14.6   &\small 29.5  &\small 28.4     &\small 29.1     &\small 10.3     &\small 13.0  &\small 25.4 &\small 31.9 &\small 30.4 &\small 25.2 \\
		\small RGB-D & \small \bf 61.2     & \small \bf 63.4     & \small \bf 44.4     & \small \bf 35.9     & \small \bf 28.2     & \small \bf 55.6     &\small \bf 63.7     &\small \bf 56.5  &\small \bf 26.9  &\small \bf 49.5     &\small \bf 47.1     &\small \bf 49.7     &\small \bf 56.2     &\small \bf 59.9     &\small \bf 46.3     &\small \bf 28.3     &\small \bf 50.4     &\small \bf 46.8 &\small \bf 55.4  &\small \bf 62.7 &\small \bf 49.4\\
		\hline
		\small \tabincell{c}{RGB \cite{girshick2014rcnn}\\(finetune)} & \small 63.5 & \small 66.0     & \small 47.9   & \small 37.7     & \small 29.9     & \small 62.5   &\small 70.2   &\small  60.2  &\small  32.0  &\small 57.9   &\small 47.0    &\small 53.5   &\small  60.1 &\small 64.2  &\small 52.2 &\small 31.3  &\small 55.0 &\small 50.0 &\small 57.7 &\small 63.0 &\small 53.1\\
		\small \tabincell{c}{RGB\\(finetune)-D} & \small \bf 67.7     & \small \bf 66.4     & \small \bf 47.8     & \small \bf 36.0     & \small \bf 31.9     & \small \bf 61.5     &\small \bf 70.6     &\small \bf 59.0  &\small \bf 32.6  &\small \bf 52.3     &\small \bf 51.4     &\small \bf 55.9     &\small \bf 61.1     &\small \bf 63.8     &\small \bf 52.8     &\small \bf 32.6     &\small \bf 57.7     &\small \bf 50.5 &\small \bf 58.1  &\small \bf 64.8 &\small \bf 54.0\\
		\hline
	\end{tabular}
\end{table*}

\begin{table*}
	\renewcommand\arraystretch{1.25}
	\newcommand{\tabincell}[2]{\begin{tabular}{@{}#1@{}}#2\end{tabular}}
	\caption{Detection results on the VOC 2007 dataset of our RGB-D  Fast R-CNN detector. The first row shows the results of RGB features only, the second row shows the results of depth features only,  and the last two rows show the result of RGB-D features.}
	\centering
	\label{table:07_estD_results_frcn}
	\begin{tabular}{@{\hskip 0.8mm}c@{\hskip 0.8mm} @{\hskip 0.8mm}c@{\hskip 0.8mm} @{\hskip 0.8mm}c@{\hskip 0.8mm} @{\hskip 0.8mm}c@{\hskip 0.8mm} @{\hskip 0.8mm}c@{\hskip 0.8mm} @{\hskip 0.8mm}c@{\hskip 0.8mm} @{\hskip 0.8mm}c@{\hskip 0.8mm} @{\hskip 0.8mm}c@{\hskip 0.8mm} @{\hskip 0.8mm}c@{\hskip 0.8mm} @{\hskip 0.8mm}c@{\hskip 0.8mm} @{\hskip 0.8mm}c@{\hskip 0.8mm} @{\hskip 0.8mm}c@{\hskip 0.8mm} @{\hskip 0.8mm}c@{\hskip 0.8mm} @{\hskip 0.8mm}c@{\hskip 0.8mm} @{\hskip 0.8mm}c@{\hskip 0.8mm} @{\hskip 0.8mm}c@{\hskip 0.8mm} @{\hskip 0.8mm}c@{\hskip 0.8mm} @{\hskip 0.8mm}c@{\hskip 0.8mm} @{\hskip 0.8mm}c@{\hskip 0.8mm} @{\hskip 0.8mm}c@{\hskip 0.8mm} @{\hskip 0.8mm}c@{\hskip 0.8mm} @{\hskip 0.8mm}c@{\hskip 0.8mm}}
		\noalign{\smallskip}
		\hline\noalign{\smallskip}
		                 &\small aero    &\small bike  &\small bird   &\small boat   &\small bottle  &\small bus &\small car    &\small cat    &\small chair   &\small cow    &\small table  &\small dog  &\small horse   &\small mbike  &\small person &\small plant &\small sheep  &\small sofa &\small train &\small tv &\small mAP \\
		\noalign{\smallskip}
		\hline
		\noalign{\smallskip}
		\small RGB  & \small 65.3 & \small 74.3  & \small 60.3   & \small 48.7     & \small 36.2   & \small 74.0   &\small 75.7   &\small 77.9  &\small 40.5 &\small   66.1 &\small 54.6   &\small 73.8    &\small 77.7    &\small 73.1     &\small 67.5     &\small 35.4    &\small 57.6   &\small 62.7 &\small 74.3 &\small 57.1 &\small 62.6\\
		\small Depth & \small 47.8  & \small 46.7     & \small 26.8     & \small 27.7     & \small 14.3     & \small 54.7    &\small 56.6  &\small 49.5  &\small 21.5  &\small 27.2 &\small 40.7   &\small 35.1   &\small 53.7  &\small 45.2   &\small 47.8   &\small 14.7   &\small 17.5  &\small 45.3 &\small 51.2 &\small 38.8 &\small 38.1 \\

		\small \tabincell{c}{RGB-D(fc6)} & \small 60.9 & \small \bf 76.7 & \small \bf 60.7   & \small 46.8    & \small 35.5   & \small 73.6  &\small \bf 76.4   &\small \bf 79.4  &\small \bf  41.5  &\small 69.8   &\small 57.3   &\small 75.5   &\small 78.2 &\small 68.7  &\small 70.6 &\small \bf 34.4  &\small \bf 60.6 &\small \bf 64.8 &\small 76.1 &\small \bf 62.6 &\small 63.5\\

		\small RGB-D(fc7) & \small \bf 65.3     & \small 75.8  & \small 59.6   & \small \bf 48.3    & \small \bf 36.8    & \small \bf 74.3    &\small 75.8    &\small  78.7  &\small 41.4 &\small \bf 70.5   &\small \bf 57.5    &\small \bf 75.9    &\small \bf 79.1    &\small \bf 68.8    &\small \bf 71.7 &\small 33.5   &\small 58.6   &\small 63.4 &\small \bf 76.6  &\small 59.7 &\small \bf 63.6\\

		\hline
	\end{tabular}
\end{table*}

\begin{table*}[!ht]
\renewcommand\arraystretch{1.2}
\newcommand{\tabincell}[2]{\begin{tabular}{@{}#1@{}}#2\end{tabular}}
	\caption{Detection results on the NYUD2 dataset. The first three columns are detection results of RGB features only. Columns 4-6 are detection results of depth features only. Columns 7-10 are detection results of combined RGB and depth features. The last two columns are detection results using features extracted from ground-truth depth. ``pool5", ``fc6", and ``fc7" denote different layers for feature extraction, ``sctratch" denotes the depth feature learning network is trained from scratch.}
	\centering
	\label{table:NYUD2_estD_results}
	\begin{tabular}{ @{\hskip 0.8mm}c@{\hskip 0.8mm} @{\hskip 0.8mm}c@{\hskip 0.8mm} @{\hskip 0.8mm}c@{\hskip 0.8mm} @{\hskip 0.8mm}c@{\hskip 0.8mm} | @{\hskip 0.8mm}c@{\hskip 0.8mm} @{\hskip 0.8mm}c@{\hskip 0.8mm} @{\hskip 0.8mm}c@{\hskip 0.8mm}  | @{\hskip 0.8mm}c@{\hskip 0.8mm} @{\hskip 0.8mm}c@{\hskip 0.8mm} @{\hskip 0.8mm}c@{\hskip 0.8mm} @{\hskip 0.8mm}c@{\hskip 0.8mm} | @{\hskip 0.8mm}c@{\hskip 0.8mm} @{\hskip 0.8mm}c@{\hskip 0.8mm}}
		\hline\noalign{\smallskip}
		      & \tabincell{c}{RGB\\(pool5)}  & \tabincell{c}{RGB \cite{guptaECCV14}\\(fc6)} & \tabincell{c}{RGB\\(fc7)} &\tabincell{c}{Depth \\(pool5)} &\tabincell{c}{Depth \\(fc6)} &\tabincell{c}{Depth \\(fc7)}  & \tabincell{c}{RGB-D \\(pool5)} & \tabincell{c}{RGB-D \\(fc6)} & \tabincell{c}{RGB-D \\(fc7)} & \tabincell{c}{RGB-D \\(fc6,scratch)} & \tabincell{c}{GT Depth} & \tabincell{c}{RGB-GTD \\(fc6)}\\
		\noalign{\smallskip}
		\hline
		\noalign{\smallskip}
		bathtub      & 16.2  & 5.5     & \bf18.3     & 0.7     & 3.9       \bf& 4.6      & 15.4     & 19.3     & \bf19.8  & 18.5  &23.9   &\bf 39.1 \\
		bed          & 38.4  & \bf52.6     & 40.7     & 35.9    & \bf40.1      & 39.3     & 51.4     & \bf52.5     & 49.1  &45.6  &58.8   &\bf 65.4\\
		bookshelf    & 22.8  & 19.5     & \bf27.4     & \bf4.4     & 4.1       & 2.0      & 26.0     & \bf30.5     & 29.0  &28.1  &27.7   &\bf 34.2\\
		box          & 0.8   & \bf1.0      & 0.8      & 0.1     & 0.1       & 0.1      & 0.8      & 0.7      \bf& 0.8      &0.6   & 0.4   &\bf  0.8\\
		chair        & 23.3  & 24.6     & \bf27.1     & 11.7    \bf& 14.6      & 13.9     & 25.3     &\bf 28.8     & 28.6  & 28.2  &31.7   &\bf 39.6\\
		counter      & 27.8  & 20.3     & \bf34.2     & 15.4    & 19.9      \bf& 20.0     & 32.4     & 36.2     &\bf 37.2  &34.9  &34.5   &\bf 45.2\\
		desk         & 5.2   & 6.7      & \bf8.8      & 3.2     & \bf4.3       & 3.4      & 8.6      & \bf11.2     & 10.4  &8.2   & 3.8   &\bf 11.8\\
		door         & 13.3  &14.1     & \bf15.0     & 1.7     & \bf1.9       & 1.9      & 13.5     & 14.1     & 15.4  &\bf15.5  & 3.6   &\bf 18.3\\
		dresser      & 10.3  &16.2     & \bf17.0     & 4.3     & 7.0       & \bf7.3      & 16.1     & \bf25.0     & 20.1  &17.1  &14.0   &\bf 25.4\\
		garbage bin  & 15.0  &\bf17.8     & 16.4     & 2.6     & \bf4.3       & 4.1      & 17.8     & \bf20.9     & 17.6  &16.5  &13.2   &\bf 28.3\\
		lamp         & 24.0  & 12.0     & \bf25.9     & 10.2    & 10.2      & \bf10.3     & 22.3     & 27.4     & \bf27.8  &26.9  &27.1   &\bf 34.1\\
		monitor      & 28.8  & 32.6     & \bf37.4     & 5.3     & \bf7.6       & 6.8      & 31.5     & 36.5     & \bf38.4  &37.1  & 7.0   &\bf 38.4\\
		nightstand   & 11.8  & \bf18.1     & 12.4     & 0.9     & 0.9       & \bf1.1      & \bf14.0     & 10.8     & 13.2  &12.2  &23.3   &\bf 31.4\\
		pillow       & 13.0  & 10.7     & \bf14.5     & 3.7     & 5.5       & \bf7.0      & 16.3     & \bf18.3     & 17.5  &16.8  &15.2   &\bf 27.1\\
		sink         & 21.2  & 6.8     & \bf25.7     & 3.6     & 7.6       & \bf9.4      & 20.0     & 26.1     & 26.3  &\bf28.1  &20.0   &\bf 36.2\\
		sofa         & \bf29.0  & 21.6     & 28.6     & 11.5    & \bf13.9      & 12.3     & 30.6     & \bf31.8     & 29.3  &27.7  &28.7   &\bf 38.2\\
		table        & 9.7   &10.0     & \bf11.2     & 7.8     & \bf10.3      & 9.6      & 14.5     & \bf15.0     & 14.4  &11.7  &21.1   &\bf 21.4\\
		television   & 23.9  & \bf31.6     & 26.0     & \bf4.4     & 3.9       & 4.3      & 27.6     & 29.3     & 28.4  &\bf29.5  & 7.8   &\bf 26.4\\
		toilet       & 40.0  &\bf52.0     & 44.8     & \bf29.5    & 27.6      & 26.9     & 42.8     & \bf44.8     & 44.2  &40.2  &42.2   &\bf 48.2\\
		mAP          & 19.7  & 19.7     & \bf22.7     & 8.3     & \bf9.9       & 9.7      & 22.5     & \bf25.2     & 24.6  &23.3  &21.3   &\bf 32.1\\
		\noalign{\smallskip}
		\hline
	\end{tabular}
\end{table*}

\subsection{RGB-D Object Detection}
In this section we show the results of our RGB-D object detection. We first show the results of our RGB-D R-CNN object detector, then we show the results of our RGB-D Fast R-CNN object detector. Finally we analyse some key components in our experiments.

\subsubsection{RGB-D R-CNN detection results}

We first show the test results on the NYUD2 dataset, which we also use to train the DCNF model. The NYUD2 dataset has 1449 RGB-D image pairs captured by a Microsoft Kinect. We use the same object proposals in \cite{guptaECCV14}. The  RGB feature learning network is also from \cite{guptaECCV14} which is fine-tuned on additional synthetic images. We use the training set to fine-tune our depth feature learning network. The detection results are shown in Table \ref{table:NYUD2_estD_results}. All the detectors are trained on the training and validation sets and tested on the test set. We choose the output of different layers (denoted as pool5, fc6, fc7 which is same with AlexNet) as depth or RGB features. As we can see from the table, with the depth information being added, the best mAP is 25.2\%, which is 5.5\% higher than the result in \cite{guptaECCV14}.

We then use the same DCNF model to estimate depths of the B3DO dataset. The B3DO dataset is a relatively smaller RGB-D dataset than the NYUD2. It consists of 849 RGB-D image pairs. The scene types of the B3DO dataset are mainly domestic and office which are very similar to the NYUD2. We generate around 2000 object proposals using selective search \cite{Uijlings13} in each image. All the detectors are trained on the training set and tested on the validation set. We follow the detection baseline in \cite{JanochKJBFSD11} and report the test results on 8 objects: chairs, monitors, cups, bottles, bowls, keyboards, computer mice and phones. But different from \cite{JanochKJBFSD11} which uses 6 different splits for evaluation and reported the averaged AP values, we only use the first split (with 306 images in the training set and 237 images in the validation set). We directly use the AlexNet without fine-tuning to extract RGB features. The results are illustrated in Table \ref{table:B3DO_estD_results}, from which we can see that the added depth features improve the detection mAP by 2.0\%.

\begin{table*}
\renewcommand\arraystretch{1.2}
	\caption{Detection results of our RGB-D R-CNN detector on B3DO dataset. The first row shows the results of RGB features only, the second row shows the results of estimated depth features only, the third row shows the results of combined RGB features and depth features.}
	\centering
	\label{table:B3DO_estD_results}
	\begin{tabular}{cccccccccc}
		\hline\noalign{\smallskip}
		                 & chair    & monitor  & cup      & bottle   & bowl     & keyboard & mouse    & phone    & mAP      \\
		\noalign{\smallskip}
		\hline
		\noalign{\smallskip}
		RGB              & 37.0     & 79.0     & 33.0     & 19.5     & 29.6     & 56.5     & 26.9     & 37.9     & 39.9     \\
		Estimated Depth  & 35.9     & 59.9     & 21.8     & 6.1      & 14.7     & 15.2     & 9.0      & 8.4      & 21.4     \\
		RGB-D (Estimated) & \bf 43.3 & \bf 78.6 & \bf 33.8 & \bf 20.2 & \bf 34.3 & \bf 57.0 & \bf 24.2 & \bf 44.0 & \bf 41.9 \\
		\hline
	\end{tabular}
\end{table*}

\begin{figure*}[t]
	\begin{center}
		\includegraphics[scale=.8]{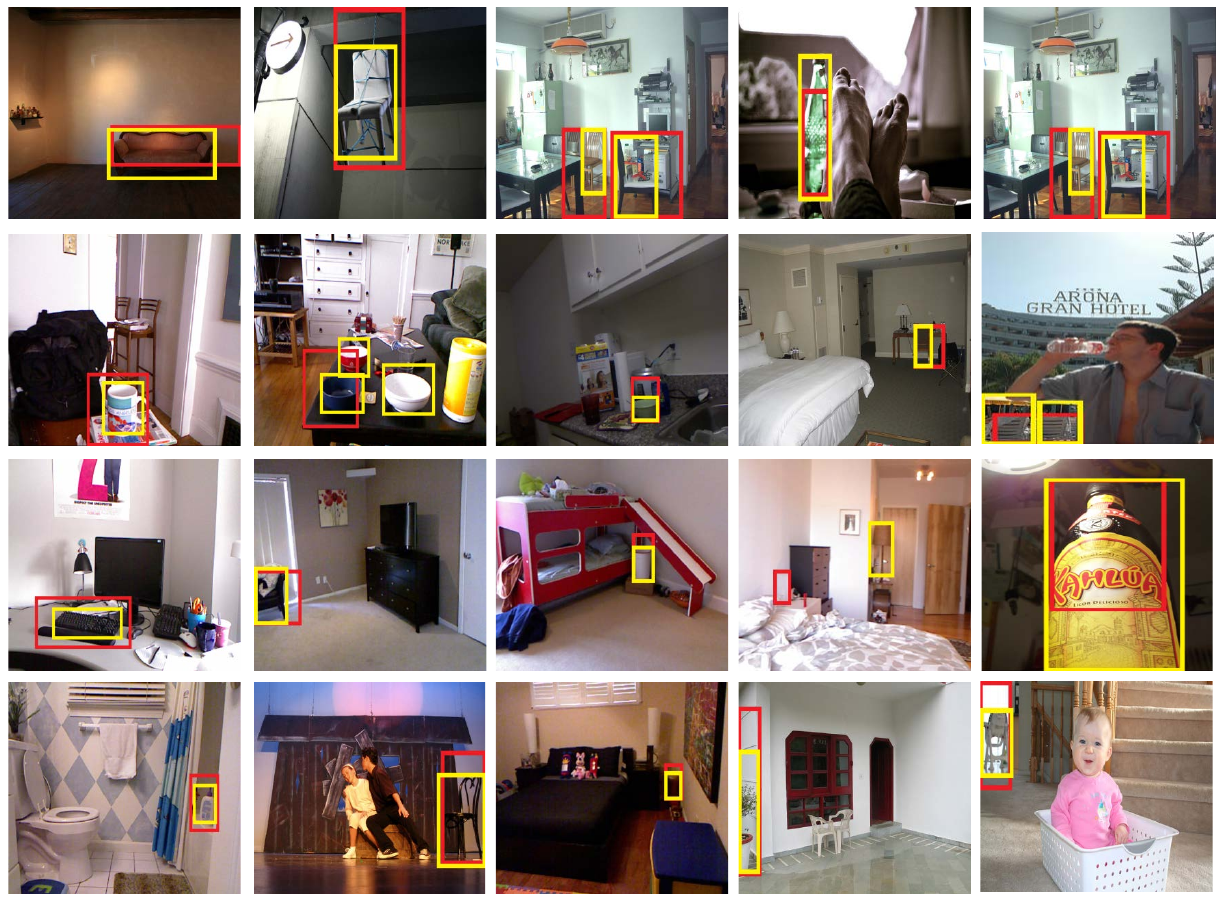}
	\end{center}
	\caption{Some detection examples. The red boxes are instances detected by the RGB detector, and the yellow boxes are instances obtained by our RGB-D detector.}
	\label{fig:examples}
\end{figure*}

\begin{table*}
	\renewcommand\arraystretch{1.5}
	\caption{Segmentation result on VOC2012 dataset. We use the augmented data for training and test on validation set. The first row shows the results using RGB images only The second row shows the result of our RGB-D segmentation by multi-task training. The last 2 rows show the results of our RGB-D segmentation by feature concatenation.}
	\centering
	\label{table:07_seg}
	\begin{tabular}{@{\hskip 0.8mm}c@{\hskip 0.8mm} @{\hskip 0.8mm}c@{\hskip 0.8mm} @{\hskip 0.8mm}c@{\hskip 0.8mm} @{\hskip 0.8mm}c@{\hskip 0.8mm} @{\hskip 0.8mm}c@{\hskip 0.8mm} @{\hskip 0.8mm}c@{\hskip 0.8mm} @{\hskip 0.8mm}c@{\hskip 0.8mm} @{\hskip 0.8mm}c@{\hskip 0.8mm} @{\hskip 0.8mm}c@{\hskip 0.8mm} @{\hskip 0.8mm}c@{\hskip 0.8mm} @{\hskip 0.8mm}c@{\hskip 0.8mm} @{\hskip 0.8mm}c@{\hskip 0.8mm} @{\hskip 0.8mm}c@{\hskip 0.8mm} @{\hskip 0.8mm}c@{\hskip 0.8mm} @{\hskip 0.8mm}c@{\hskip 0.8mm} @{\hskip 0.8mm}c@{\hskip 0.8mm} @{\hskip 0.8mm}c@{\hskip 0.8mm} @{\hskip 0.8mm}c@{\hskip 0.8mm} @{\hskip 0.8mm}c@{\hskip 0.8mm} @{\hskip 0.8mm}c@{\hskip 0.8mm} @{\hskip 0.8mm}c@{\hskip 0.8mm} @{\hskip 0.8mm}c@{\hskip 0.8mm}}
		\noalign{\smallskip}
		\hline\noalign{\smallskip}
		                 &\small aero    &\small bike  &\small bird   &\small boat   &\small bottle  &\small bus &\small car    &\small cat    &\small chair   &\small cow    &\small table  &\small dog  &\small horse   &\small mbike  &\small person &\small plant &\small sheep  &\small sofa &\small train &\small tv &\small mean \\
		\noalign{\smallskip}
		\hline
		\noalign{\smallskip}
		\small RGB  & \small 69.2  & \small \bf 31.0   & \small 64.5     & \small 48.6     & \small 49.9   &\small \bf 73.2   &\small 68.3 &\small 72.2 &\small \bf 24.2 &\small 50.9    &\small 34.2  &\small 61.2  &\small 52.9   &\small \bf 64.4  &\small \bf 72.0     &\small \bf 41.7    &\small 59.1 &\small 27.5 &\small 67.9 & \small 57.7  &\small 56.2\\

		\small RGB-D\\(multi-task) & \small \bf 69.5     & \small  30.2     & \small \bf 68.4     & \small \bf 53.3     & \small \bf 54.1     & \small  72.8     &\small \bf 70.3    &\small \bf 72.2  &\small  22.4     &\small \bf 54.6     &\small \bf 36.5     &\small \bf 63.5     &\small \bf 55.9     &\small  63.7     &\small 71.5     &\small  40.7      &\small \bf 60.2     &\small \bf 31.5 &\small \bf 69.7  &\small \bf 59.5 &\small \bf 57.6\\
		\hline

		\small RGB-D(fc6)  & \small \bf 70.9  & \small \bf 29.9   & \small \bf 68.5    & \small 54.9   & \small \bf 55.8   &\small \bf 75.0   &\small \bf 72.0 &\small \bf 72.6 &\small 21.2 &\small \bf 53.5   &\small \bf 37.5  &\small 63.4  &\small \bf 56.2  &\small 63.4  &\small \bf 72.6  &\small \bf 43.7  &\small \bf 61.6 &\small 32.6 &\small \bf 71.2 & \small \bf 60.0 &\small \bf 58.4\\

		\small RGB-D(fc7)  & \small 69.4  & \small 29.9   & \small 68.3    & \small \bf 55.7   & \small 53.5   &\small 73.2   &\small 71.4 &\small 72.4 &\small \bf 22.6 &\small 53.1   &\small 37.0  &\small \bf 63.9  &\small 55.3  &\small \bf 63.7  &\small 71.7  &\small 42.7  &\small 61.6 &\small \bf 33.2 &\small 70.4 & \small 57.7 &\small 57.9\\
		\hline
	\end{tabular}
\end{table*}

Both the NYUD2 and B3DO  are RGB-D datasets captured by the Microsoft Kinect and have similar scenes. In order to further show the the effectiveness of our deeply learned depth feature, we test object detection on the PASCAL VOC 2007 and 2012 datasets which have no measured depths at all. We use the same regional proposals in \cite{girshick2014rcnn} which are also generated by selective search. We first report the results on indoor objects in the VOC2012 dataset in Table \ref{table:VOC12_estD_results}. There are 6 indoor objects among all the 20 objects in the VOC2012 dataset, including bottle, chair, dining table, potted plant, sofa, tv/monitor. We select all the images containing these 6 objects and use the training set (1508 indoor images) to train the depth feature learning network and detector, and use the validation set (1523 indoor images) for testing. We directly use the AlexNet without finetuning for RGB feature extraction and use the output of the ``fc6" layer as feature vectors. As we can see from Table \ref{table:VOC12_estD_results}, our learned depth features improve the detection mAP by 3.6\%.

Table \ref{table:07_estD_results} shows the detection results on the VOC2007 dataset. We train our depth feature learning network and SVMs on training and validation sets and report results on test set. Since we use all images in the VOC2007 dataset, we train the DCNF model using a combination of the NYUD2 and Make3D datasets. We first directly use AlexNet without fine-tuning for the RGB feature extraction. As we can see from the first 3 rows, the additional depth features improve the mAP by 3.2\%. We then use fine-tuned  AlexNet for the RGB feature extraction and report the results in the last two rows. As we can see, even though the RGB fine-tuning increases correlativity between color and depth information, the additional depth feature still improves the detection mAP by 0.9\%. Similar to the experiment on the VOC2012 dataset, we use the output of the ``fc6" layer as the RGB and depth features. Some examples of our RGB-D detection are shown in Fig. \ref{fig:examples}.

\begin{table*}
	\renewcommand\arraystretch{1.5}
	\caption{Segmentation result of our RGB-D segmentation by feature concatenation on VOC2012 dataset. We use the standard 1464 images for training and test on validation set. The first row shows the results using RGB features only, and the last 3 row shows the results with additional depth features.}
	\centering
	\label{table:std_12_cat_seg}
	\begin{tabular}{@{\hskip 0.8mm}c@{\hskip 0.8mm} @{\hskip 0.8mm}c@{\hskip 0.8mm} @{\hskip 0.8mm}c@{\hskip 0.8mm} @{\hskip 0.8mm}c@{\hskip 0.8mm} @{\hskip 0.8mm}c@{\hskip 0.8mm} @{\hskip 0.8mm}c@{\hskip 0.8mm} @{\hskip 0.8mm}c@{\hskip 0.8mm} @{\hskip 0.8mm}c@{\hskip 0.8mm} @{\hskip 0.8mm}c@{\hskip 0.8mm} @{\hskip 0.8mm}c@{\hskip 0.8mm} @{\hskip 0.8mm}c@{\hskip 0.8mm} @{\hskip 0.8mm}c@{\hskip 0.8mm} @{\hskip 0.8mm}c@{\hskip 0.8mm} @{\hskip 0.8mm}c@{\hskip 0.8mm} @{\hskip 0.8mm}c@{\hskip 0.8mm} @{\hskip 0.8mm}c@{\hskip 0.8mm} @{\hskip 0.8mm}c@{\hskip 0.8mm} @{\hskip 0.8mm}c@{\hskip 0.8mm} @{\hskip 0.8mm}c@{\hskip 0.8mm} @{\hskip 0.8mm}c@{\hskip 0.8mm} @{\hskip 0.8mm}c@{\hskip 0.8mm} @{\hskip 0.8mm}c@{\hskip 0.8mm}}
		\noalign{\smallskip}
		\hline\noalign{\smallskip}
		                 &\small aero    &\small bike  &\small bird   &\small boat   &\small bottle  &\small bus &\small car    &\small cat    &\small chair   &\small cow    &\small table  &\small dog  &\small horse   &\small mbike  &\small person &\small plant &\small sheep  &\small sofa &\small train &\small tv &\small mean \\
		\noalign{\smallskip}
		\hline
		\noalign{\smallskip}
		\small RGB  & \small 52.3  & \small 23.2   & \small 41.8    & \small 34.5     & \small 36.9   &\small 60.4   &\small 56.3 &\small 53.4 &\small 11.9 &\small 14.4    &\small 32.5  &\small 44.5  &\small 33.4  &\small 50.3  &\small 63.1   &\small 23.4    &\small 41.4 &\small 14.7 &\small 48.2 & \small 45.0  &\small 41.4\\

		\small RGB-D (pool5) & \small 57.7     & \small  15.5     & \small 50.5   & \small 38.8    & \small 41.6   & \small 60.8    &\small 58.1    &\small 59.1  &\small 13.9   &\small 33.6    &\small 29.9    &\small 48.7    &\small 33.3   &\small 52.0    &\small 65.0   &\small  27.9   &\small 45.4   &\small 26.4 &\small 52.6  &\small 43.3 &\small 44.8\\
		\small RGB-D (fc6) & \small \bf 61.8    & \small \bf 25.7    & \small \bf 58.5  & \small \bf 47.5   & \small \bf 49.7   & \small \bf 67.3  &\small 65.7  &\small \bf 65.1  &\small 15.6 &\small 42.5  &\small 36.5  &\small \bf 55.9   &\small \bf 44.9  &\small \bf 58.1   &\small \bf 67.0  &\small \bf 34.2  &\small 54.0  &\small \bf 31.2 &\small \bf 60.8  &\small \bf 48.5 &\small \bf 51.4\\
		\small RGB-D (fc7) & \small 61.5    & \small  24.0    & \small 57.1  & \small 44.8   & \small 47.9   & \small 66.9  &\small \bf 66.1  &\small 63.9  &\small 16.0 &\small \bf 43.1  &\small \bf 36.8  &\small 54.8   &\small 44.0  &\small 56.7   &\small 66.8  &\small  33.0  &\small \bf 54.3  &\small 29.7 &\small 60.2  &\small 47.6 &\small 50.6\\
		\hline
	\end{tabular}
\end{table*}

\begin{table}
	\renewcommand\arraystretch{1.2}
	\caption{The VOC2012 segmentation IoU score of our method on the \textit{validation} set using the standard \textit{training} set for training. The coefficient $\lambda$ is a trade-off multiplied by the depth error in the back-propagation. $n$ is the number of fully-connected layers in the depth processing stream.}
	\centering
	\label{table:07_seg_std}
	\begin{tabular}{ @{\hskip 0.9mm}c@{\hskip 0.9mm} @{\hskip 0.9mm}c@{\hskip 0.9mm} @{\hskip 0.9mm}c@{\hskip 0.9mm} @{\hskip 0.9mm}c@{\hskip 0.9mm} @{\hskip 0.9mm}c@{\hskip 0.9mm} @{\hskip 0.9mm}c@{\hskip 0.9mm}}
		\noalign{\smallskip}
		\hline
		\noalign{\smallskip}
		    &  $\lambda=0$ &  $\lambda=0.1$  & $\lambda=1$ & $\lambda=10$ & $\lambda=50$\\
		 \hline
		 \noalign{\smallskip}
		 $n=2$ &  41.4  &  50.2    &  49.9   &  49.1  & 46.8 \\
         \hline
         \noalign{\smallskip}
		 $n=3$ &   - & 50.0 & 50.2  &  49.4     &  49.8\\
		\hline
		\noalign{\smallskip}
		 $n=5$ &   -   & 50.0 &  \bf 50.4     &  49.9     &  48.3\\
		 \hline
	\end{tabular}
\end{table}

\subsubsection{RGB-D  Fast R-CNN detection results}
We test our RGB-D Fast R-CNN detection on VOC2007 dataset and show the results on Table \ref{table:07_estD_results_frcn}. We use the object proposals generated by RPN network in \cite{ren2015faster}. Similar to \cite{ren2015faster}, we use 2000 RPN proposals in each image to train and test on 300 RPN proposals in each image. In order to fit the GPU memory, we use single-scale training $s=300$ pixels and cap the longest image side at 500 pixels (600 and 1000 respectively in \cite{girshickICCV15fastrcnn}). We also use a mini-batch size of 1 and only sample 32 RoIs from one image. The experiments in \cite{girshickICCV15fastrcnn} use a mini-batch size of 2 and sample 128 RoIs from 2 images. As we can see from Table \ref{table:07_estD_results_frcn}, our RGB-D detection mAP outperforms RGB only detection mAP by 1.0\%. Notably, our RGB only detection results is lower than \cite{ren2015faster}. This is mainly because we shrink the input image size and especially the mini-batch size as the RoIs in one image are correlated.

Additionally, we can also learn an RPN network with our estimated depth incorporated. Specifically, we can apply a two steam structure with RGB and depth images as two separate inputs. After two processing streams that do not share parameters, the RGB and depth feature maps are concatenated to generate object proposals. With the estimated depth being incorporated in RPN network learning, our RGB-D detection can also be applied on the latest Faster-RCNN.

\subsubsection{Ceiling performance}
Since we use estimated depth for depth feature extraction, we need to find out is there any space for further improvement of detection by improving depth estimation accuracy. We test our RGB-D R-CNN object detection using ground-truth depth on NYUD2 dataset and show the results in the last two columns in Table \ref{table:NYUD2_estD_results}. As we can see from Table \ref{table:NYUD2_estD_results}, the detection mAP of ground-truth depth only is 21.3\%, which is even better than the RGB only mAP. And the mAP by ground truth RGB-D outperforms the mAP by estimated RGB-D by 6.9\%. From this experiment we can see that depth estimation is a key component of our RGB-D detection methods, the detection performance can be further improved by improving depth estimation accuracy.

\subsubsection{Network initialization}
In the aforementioned RGB-D R-CNN experiments, we initialize our depth feature learning network with AlexNet and fine-tune it using depth images based on the similarities among RGB and depth images. We also duplicate depth map to 3 channels to fit the input of the pre-trained AlexNet. In order to show the effectiveness of this step, we train a depth feature learning network from scratch on a larger RGB-D dataset and test object detection on NYUD2 dataset. Specifically, we use SUNRGBD \cite{Song_2015_CVPR} which contains a total number of 10335 images. Since we report detection result on NYUD2 dataset, we use all the images exclude the 654 NYUD2 test images. We use selective search to generate around 2000 object proposals in each image. We show the results in Table \ref{table:NYUD2_estD_results} denoted as RGB-D (fc6,scratch). As we can see from the result, AlexNet initialization works better than training depth learning network from scratch. This experiment confirms that there are certain similarities between RGB and depth images and we can take advantage of networks pre-trained on RGB datasets to exploit depth information.

In addition, we also replace the depth features with RGB features learned from another network. Specifically,  we initialize another RGB feature learning network with Place-AlexNet \cite{NIPS2014_5349} which is trained on Place dataset with 2.5 million images. We fine-tune this network also with RGB images to extract new RGB features (output of fc6). Then all the two streams in are RGB features. We test detection with the two RGB features and get an mAP of 21.9\%. This experiment further confirms that the information exploited from depth images is useful for RGB only tasks.

\subsection{RGB-D Semantic Segmentation}
In this section, we first show the result of our RGB-D segmentation by feature concatenation, then we show the result of our RGB-D segmentation by multi-task training. All the segmentation results are reported on the validation of VOC2012 dataset which contains 1449 images. The standard segmentation training set of VOC2012 has 1464 images. Following the conventional setting in \cite{BharathECCV2014} and \cite{ChenPKMY14}, we also report the segmentation results using the 10582 training images augmented in \cite{BharathICCV2011}.

\subsubsection{RGB-D segmentation by feature concatenation}
We first train our RGB-D segmentation network using the standard 1464 images and show the result in Table \ref{table:std_12_cat_seg}. Similar to our RGB-D detection experiments, we also concatenate the RGB and depth features at different places (denoted as pool5, fc6,fc7 which is same with VGG16 net) and report the results. As we can see from the table, all the results with depth features outperform the RGB only result. And the maximum mean IoU score improvement is 10\% when we choose the output of fc6 layer to be the depth feature.

Next we train our RGB-D segmentation network using augmented images and report the result in Table \ref{table:07_seg}. From the table we can see that the maximum mean IoU score improvement is 2.2\%, which is much less comparing to the improvement by standard training data. This is caused by the high correlativity between the estimated depth maps and the RGB images. When the RGB training images are sufficient, there is less additional information the depth maps can provide.

\subsubsection{RGB-D segmentation by multi-task training}
In our RGB-D segmentation network by feature concatenation, the separate feature extraction networks minimize the correlativity between the estimated depths and the RGB images. While in our RGB-D segmentation network by multi-task training, since the depth and color information share parameters, it is a question how much the estimated depth can help in updating the network.

During back-propagation, there are two factors controlling the extent of the participation of depth information in updating the network parameters. The first one is a trade-off hyper-parameter $\lambda$ multiplied by the lease squared loss. When $\lambda$ is 0, the depth information contributes nothing to the parameter updating. The second one is the number of convolution layers $n$ in the depth information processing stream. Fig. \ref{fig:rgbd-seg} shows an example of $n=5$. We also test $n=2$ and $n=3$. When $n=2$, the depth information processing stream contains 2 convolution layers with 128 and 1 channels respectively. When $n=3$, the depth information processing stream contains 5 convolution layers with channels 256, 128 and 1 respectively. This factor controls the number of common layers that shared by color and depth information in the network.

We first use the standard 1464 images for training and report the test results in Table \ref{table:07_seg_std}. As we can see from the table, the maximum improvement introduced by depth information is 9\%. We also notice that smaller participation of depth information in updating network parameters (larger value of $n$ and smaller value of $\lambda$) lead to better performance. This is because the smaller participation of depth information, the less correlativity between the depths and RGB images.

Combining losses is often hard to tease apart from adjusting the learning rate or other regularizations, since adding a loss effectively changes the gradient size. In  the aforementioned experiments, we initialize the learning rate (lr) to be $1e^{-7}$ and decrease by a factor of 0.4 every 5 epochs. The weight decay (wd) is set to 0.0005. In order to make sure the performance improvement is introduced by the exploited depth information, we also test on 5 other different learning rate and weight decay schemes: (a) $wd=0.001$, initial $lr=1e^{-7}$ and decrease 0.4 every 5 epochs, (b) $wd=0.0001$, initial $lr=1e^{-7}$ and decrease 0.4 every 5 epochs,(c) $wd=0.0005$, initial $lr=1e^{-7}$ and decrease 0.1 every 5 epochs, (d) $wd=0.0005$, initial $lr=1e^{-7}$ and decrease 0.4 every 7 epochs, (e) $wd=0.0005$, initial $lr=1e^{-8}$ and decrease 0.4 every 5 epochs. We get the mean IoU score of 50.2\%, 50.3\%, 50.4\%, 50.1\% and 48.8\% respectively. These results confirm that the performance improvement is introduced by exploited depth information.

Next we we show the results using augmented training images in Table \ref{table:07_seg}. We set $\lambda=10$ and $n=3$. As we can see, the mean IoU score is only improved by 1.4\% when incorporating depth information. Similar to the result of our RGB-D segmentation by feature concatenation, the improvement is much less when data augmentation is applied due to the limited additional information provided by the estimated depth.

\section{Conclusion}\label{chapt7}
We have combined the task of depth estimation with object detection and semantic segmentation and proposed two ways of exploiting depth information from estimated depth to improve the performance of object detection and semantic segmentation. Experiment results show the effectiveness of our proposed methods.

We have explored the question of whether separately estimating depths from RGB images and incorporating them as a cue for detection and segmentation improves performance, and shown that it does.  This holds despite the fact that the depth estimation process only has access to the same test data as the main (detection or segmentation) algorithm. This is particularly interesting because it shows that it is possible to improve performance by exploiting related data which does not share the same set of labels.

\section*{Acknowledgment}

This work is in part supported by ARC grant FT120100969.

\ifCLASSOPTIONcaptionsoff
\newpage
\fi

\bibliographystyle{IEEEtran}
\bibliography{CSRef}
\end{document}